\documentclass[runningheads]{llncs}
\usepackage[T1]{fontenc}
\usepackage{graphicx,subcaption,ragged2e}
\newcommand{\repeatthanks}{\textsuperscript{\thefootnote}}
\begin{document}
\title{Child Speech Recognition in Human-Robot Interaction: Problem Solved?}
%
%
\author{Ruben Janssens\thanks{Equal contribution and joint first authors.} 
\and Eva Verhelst\repeatthanks 
\and Giulio Antonio Abbo 
\and \\ Qiaoqiao Ren 
\and Maria Jose Pinto Bernal 
\and Tony Belpaeme 
}
\authorrunning{R. Janssens and E. Verhelst et al.}
%
\institute{IDLab-AIRO, Ghent University -- imec, Ghent, Belgium
\email{\{firstname.lastname\}@ugent.be}}

\maketitle              
\begin{abstract}
Automated Speech Recognition shows superhuman performance for adult English speech on a range of benchmarks, but disappoints when fed children's speech. This has long sat in the way of child-robot interaction. Recent evolutions in data-driven speech recognition, including the availability of Transformer architectures and unprecedented volumes of training data, might mean a breakthrough for child speech recognition and social robot applications aimed at children. We revisit a study on child speech recognition from 2017 and show that indeed performance has increased, with newcomer OpenAI Whisper doing markedly better than leading commercial cloud services. Performance improves even more in highly structured interactions when priming models with specific phrases. While transcription is not perfect yet, the 
best model recognises 60.3\% of sentences correctly barring small grammatical differences, with sub-second transcription time running on a local GPU, showing potential for usable autonomous child-robot speech interactions.

\end{abstract}
\section{Introduction and Background}

Spoken language interaction is for many the holy grail in HCI and HRI. It is built upon a collection of technologies, such as Automated Speech Recognition, Dialogue Management, and Text-to-Speech, that are chained together to create a system which allows the user to interact or converse with an artificial system using the most natural interface known to humankind. While this processing chain is brittle, the point of entry is Automated Speech Recognition (ASR). The ability to automatically transcribe speech utterances ---converting continuous acoustic signals into discrete symbolic representations, typically text--- has been studied extensively in academic and industrial research. In recent decades, ASR performance has come along in leaps and bounds, with companies claiming ``super-human performance'' on conversational ASR benchmarks in 2017 \cite{xiong2018microsoft}. On certain benchmarks and for resource-rich languages, both in terms of training data availability and priorities imposed by economic returns on investment, speech recognition performance is on par or even better than mean human transcription performance. The popular metric for ASR performance is Word Error Rate (WER), calculated as the total number of errors ---substitutions, insertions, and deletions--- divided by the total number of words in the text \cite{kuhn2024measuring}. WER was typically reported to be below 5\%. These systems relied on neural networks such as CNNs and LSTMs to extract features from audio signals and convert time series to text. Combined with large, annotated training sets and unsupervised learning, these systems improved over earlier model-based learning. However, while impressive, these systems' performance degraded catastrophically on speech for which it was not optimised, including atypical voices such as the speech of elderly or young children. This has repercussions for HRI and specifically for applications in which autonomous social robots are expected to interact with non-typical users, such as robots for elder care or robots for education \cite{belpaeme2018social,verhelst2024adaptive,cifuentes2020social}. 

In 2017, Kennedy \textit{et al.} \cite{kennedy2017child} published a widely cited study showing that then state-of-the-art ASR could not reliably transcribe the speech of 5-year-old English speakers. They recorded speech from 11 children in a primary school in the U.K. The speech ranged from constrained utterances ---such as counting from 1 to 10--- to unconstrained telling of a story from a picture book. Recordings were made using three different microphones, to evaluate whether the quality and hardware integration of the microphone into a robot had an impact on ASR. The ASR performance was evaluated for four different engines, three commercial ASR solutions ---Nuance VoCon 4.7, Microsoft Speech API (2016), Google Speech API (2016)--- and CMU PocketSphinx, the leading open-source solution at the time. The results were nothing but disappointing. While WER for adult speech was below 5\%, most engines could not correctly transcribe a single child utterance. Only Google's ASR did marginally better, recognising 11.8\% of constrained child speech and about 6\% of spontaneous child speech. Still, only correctly being able to transcribe 1 utterance out of 10 is a recipe for interaction disaster, and the authors of the study at the time recommended against relying on ASR for child-robot interaction.

Forward 6 years. Artificial intelligence has been revolutionised by the Transformer architecture for sequence-to-sequence tasks, not only resulting in a sea of change in the performance of generative language models but also in the performance of ASR \cite{latif2023transformers}. In September 2022, OpenAI released Whisper, an ASR engine built using an encoder-decoder Transformer architecture trained on an unprecedented 680,000 hours of labelled audio data \cite{radford2023robust}. While the specifics of Whisper's training regimen and its training data are proprietary to OpenAI, the inference model is released as public open-source software. Whisper's performance on average is better than competing solutions, but was found to still be subpar to solutions that have been specifically trained or fine-tuned on specific datasets, such as LibriSpeech \cite{radford2023robust}.

Next to the publicly available Whisper models, which still require one to install and run the ASR on own hardware, there are several cloud-based solutions. In this area large players ---Amazon, Google, Microsoft and Tencent--- compete with smaller, sometimes specialised vendors, but all offer convenient online services that are easily integrated within code. 

Given the availability of new architectures trained on larger and more diverse corpora, the time is opportune to revisit the results from Kennedy \textit{et al.} \cite{kennedy2017child} and evaluate whether state-of-the-art ASR can now handle child speech. 

We decided to compare OpenAI's Whisper, as it is open-source and exemplifies the new direction in data-driven ASR, and two commercial cloud-based solutions, opting for Microsoft Azure Speech to Text, due to its popularity and the fact that we integrate it into our robot systems at Ghent University, and Google Cloud Speech-to-text. We are first and foremost interested in transcription accuracy, but for our aim of integrating child speech recognition into an interactive HRI scenario, we also wish to explore how responsive different systems are and to which extent they would support real-time spoken interaction.

\section{Methodology}


To evaluate the ASR engines, we use the data from \cite{kennedy_2016_200495} which contains audio recordings (44KHz lossless WAV files) of 11 young children (age M=4.9 years old; 5 females, 6 males) recorded at an English primary school. The recordings consist of spontaneous speech (retelling a picture book, ‘Frog, Where Are You?’ by Mercer Mayer) and speech in which children count from 1 to 10 or repeat short sentences spoken by an adult (such as ``the horse is in the stable''). Each sample is recorded from 3 sources: a studio-grade microphone (Rode NT1-A), a portable microphone (Zoom H1) and the two front microphones of the Aldebaran NAO robot. All recordings have been manually transcribed and this is used as ground truth.


We evaluated three ASR engines: Microsoft Azure Speech to text, Google Cloud Speech-to-text, and OpenAI's Whisper. The Azure and Google models were used through a cloud API.
Whisper exists in different model sizes: tiny (39M parameters), base, small, medium, and large (1550M parameters), with three versions of the large model.
All seven of these models are compared in this study: we expect the smaller models to run faster but have lower accuracy. We used the \texttt{faster-whisper}\footnote{github.com/SYSTRAN/faster-whisper} reimplementation of the Whisper models, which claims a transcription time of up to four times faster than OpenAI's original Whisper implementation.
The Whisper models were run locally on an NVIDIA GeForce GTX 1080 Ti with 11 GB of VRAM. We also ran them on only a CPU, to assess the necessity of a dedicated GPU for these models.
We configured the models to expect English language speech, as preliminary testing revealed that without this option Whisper Large-v3 correctly detected English in only 84\% of spontaneous speech samples. All transcriptions were performed in 2024.

The performance of the models is compared using four different metrics, to ensure comparability with the results reported in \cite{kennedy2017child}.
Primarily, we use the Levenshtein distance at the letter level, which represents the minimum amount of insertions, deletions and substitutions required to change one sequence into the other. Using this metric, small errors are penalised less than they would be when using a metric like the Word Error Rate. For example, when the word ``robots'' is recognized instead of the word ``robot'', the Levenshtein distance would be 1 (as only one edit is needed to change the recognised word into the original word). We then normalise this metric by the amount of letters in the ground truth sequence. A score of 0 means perfect recognition, a score of 1 could reflect a recognised sequence of the same length but with no single letter in the right position.
Furthermore, we also report the recognition percentage, which represents the amount of utterances that are completely correctly recognised. Also, a relaxed accuracy is reported: this measure counts how many utterances are correctly recognised, also counting as accurate those with small grammatical differences that do not impact the meaning of the utterance, following the same rules as in \cite{kennedy2017child}.
Finally, we report the WER, for comparability with other ASR research, where it is the most often used metric \cite{kuhn2024measuring}.

To estimate the possibility of real-time interactions, we explore the responsiveness of the different systems by reporting their transcription time. For all Whisper models, the transcription time is the time it takes for the model to return a result, which varies due to the model size as well as the hardware on which it runs. As the Azure and Google systems are cloud-based, their transcription time also includes the transmission time of the audio file and the result
.

In all analyses, unless otherwise stated, we use only the studio microphone recordings, and only the recordings of the sentences that the children repeat from the adult ($n=50$) and of the spontaneous utterances (split into sentences, $n=222$), because preliminary analysis showed that utterances consisting of a single number are often too short for the engines to detect any speech.

\section{Results}

We will first compare the models' transcription accuracy, also investigating the impact of the length of the utterance, and improving performance by using model priming techniques. Then, we report the models' responsiveness, the impact of the microphone used, and finally a reflection on the power consumption of the models.

\subsection{Transcription accuracy}

\begin{figure}[t]
    \centering
    \includegraphics[width=0.5\columnwidth]{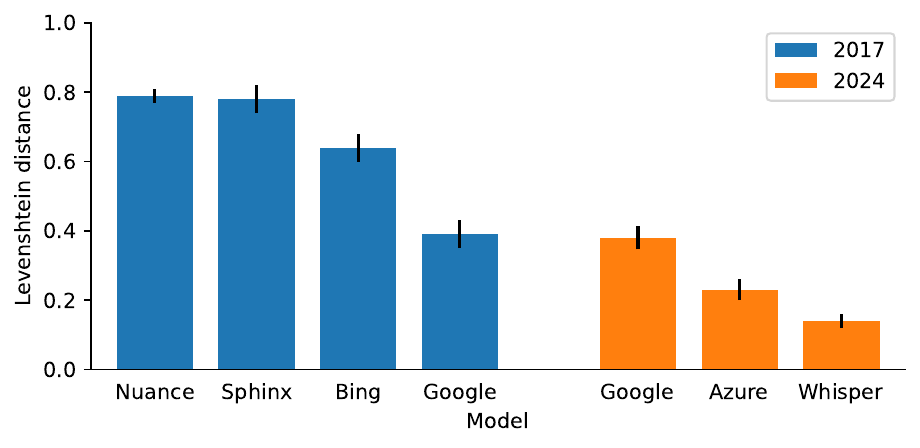}
    \caption{Performance of ASR engines in 2017 and 2024, calculated as mean normalised Levenshtein distance between ground truth and transcription (lower is better).}
    \label{fig:Levenshtein_new_old}
    \vspace{-4mm}
\end{figure}

First of all, we compare the performance of Google, Azure and the best Whisper model (large-v3) with the four engines reported in the 2017 paper. These results are shown in Figure \ref{fig:Levenshtein_new_old}. They show that the Google speech recognition did not improve compared to 2017 (Levenshtein distance $LD = 0.38$ in 2017 and in 2024), but the performance of both the Azure model ($LD = 0.23$), and the Whisper model  ($LD = 0.14$) are better than all models tested in the 2017 paper, with Whisper performing best of all.

This is also reflected in the recognition percentage: in 2017, Google was able to recognise 7.5\% of utterances correctly, in 2024, this became 9.6\%, Azure recognises 23.5\%, and Whisper 36.8\%.

The relaxed accuracy score gives an impression of the usability of the models: in 2017, Google recognised 20.3\% of the utterances correctly using relaxed criteria. This was only 14.7\% in 2024, with Azure achieving 43.0\% and Whisper 60.3\%.

These results translate into a WER of 49.0\% for Google, 30.3\% for Azure and 21.3\% for Whisper. WER was not reported in the 2017 paper. While these numbers lead to the same conclusions as the Levenshtein distance, they also show that there is still a performance gap with adult speech.

While this is not yet an ideal performance level, the relaxed accuracy shows that Whisper is already rather usable, as small mistakes that do not count as accurate for the relaxed accuracy criteria, could still be handled by dialogue management software. Table \ref{tab:examples} shows some examples of small mistakes still made by Google, Azure and the best-performing Whisper model.


Figure \ref{fig:Levenshtein_models} shows a more detailed comparison between the different model sizes of Whisper and the Azure and Google services. As expected, the large Whisper models perform best, with Whisper large v3 performing best of all.

The Kruskal-Wallis test reveals significant differences between the Levenshtein distance for the tested models ($p < 0.001$), and post-hoc Dunn tests with Bonferroni correction do not show significant differences between Whisper large v3 and Whisper small, but do show a significant difference between, among others, all large Whisper models and Azure ($p < 0.005$), and between Azure and Google ($p < 0.001$).

\begin{table}
    \centering
    \caption{Examples of small transcription mistakes}
    \begin{tabular}{ll}
        \textbf{Ground truth} & the dog is in front of the horse \\
        \hline
        \textbf{Whisper} & the dog is the front of the horse \\
        \textbf{Azure} & the dog is the front of the horse \\
        \textbf{Google} & the song in the front of the horse\\
    \end{tabular}
    \label{tab:examples}
\end{table}


\subsection{Utterance length}


The previous section reported performance on the samples containing sentences that the children repeated from an adult, and spontaneous utterances that were split into sentences. However, the dataset also contained samples containing a number that was repeated from the adult, as well as full-length spontaneous utterances. These samples are respectively much shorter (mean length 0.98s, s.d. 0.43s) and much longer (108.91s, s.d. 69.6s) than the repeated sentences (2.85s, s.d. 0.61s) and free speech cut into sentences (4.20s, s.d. 2.09s).

As shown in Table \ref{tab:utterance_types}, the ASR systems perform considerably worse on these shorter and longer samples. Only half of the repeated numbers are correctly recognized.
Looking at the mistakes the models made shows that the samples containing a number are often too short to detect any speech. On the other hand, the samples containing the full spontaneous speech fragments contain pauses between sentences, which sometimes lead to early termination of the transcription.

\vspace{-0.6cm}
\begin{table}
    \centering
\caption{Levenshtein distance for different utterance types (accuracy in brackets).}
\label{tab:utterance_types}
    \begin{tabular}{lccc}
         &  \textbf{Google} &  \textbf{Azure} & \textbf{Whisper}\\
         \hline
         \textbf{Repeated sentences}&  0.26&  0.16& 0.13\\
         \textbf{Free speech sentences}&  0.41&  0.25& 0.14\\
         \textbf{Repeated numbers}&  0.80 (17\%)&  0.38 (52\%)& 0.50 (48\%)\\
         \textbf{Full free speech samples}&  0.43&  0.89& 0.55\\
    \end{tabular}
\end{table}
\vspace{-0.9cm}

\subsection{Model priming}


All three ASR systems we evaluated offer functionalities to prime the models to expect certain words or phrases. In a setting as in this dataset, where the context of the conversation is known, this could improve performance.

For the repeated numbers, we primed the Google and Azure ASR systems to expect a number -- specifically between one and ten for Azure. Whisper works differently: it can be provided with a prompt, which acts as conversational context, which the model will try to complete with the transcript. For the numbers, the prompt \textit{``I choose number''} worked best. As shown in Table \ref{tab:prompted_results}, model priming strongly improves performance for these samples. Interestingly, the Whisper medium model performs better than large v3, reaching an average Levenshtein distance of 0.17 and an absolute accuracy of 88.0\%.



For the repeated sentences, we compare providing the five full sentences that the children said, making this a multiple-choice task, providing a template grammar, as also described in \cite{kennedy2017child}, and providing the separate words and subphrases that are part of the template. For Whisper, the template is provided as \textit{``Choose the words that were said by this child: "The <dog/fish/horse> is <in/next to/in front of/behind/on top of> the <pond/shed/car/stable/horse>"''}. Instead of providing separate words to Whisper, the prompt \textit{``Where is the animal?''} is used. Table \ref{tab:prompted_results} also shows improved performance. As expected, the more structured the priming becomes, the better the performance: Whisper achieves almost perfect recognition in the multiple-choice setting.

For the spontaneous speech samples, Google and Azure were primed with the animals that were part of the template, while Whisper was prompted with \textit{``Today we'll read `Frog, where are you?'. This book chronicles the humorous adventures that befall a young boy as he and his dog search for the frog that escaped from a jar in his room. Let's begin.}'' Here, the performance gain is more limited. For Whisper, this translates into a WER for the sponteneous sentences of 21.9\% with the prompt compared to 23.1\% without.

\vspace{-0.5cm}
\begin{table}
    \centering
\caption{Levenshtein distance for primed models (accuracy between brackets).}
\label{tab:prompted_results}
    \begin{tabular}{lllll}
          \textbf{Utterance type} &\textbf{Prompt}&  \textbf{Google}&  \textbf{Azure}& \textbf{Whisper}\\
          \hline 
         \textbf{Numbers}&None&  0.80 (17\%)&  0.38 (52\%)& 0.77 (40\%)\\
          &Number&  0.66 (33\%)&  0.32 (64\%)& 0.53 (72\%)\\
          \textbf{Repeated sentences}&None&  0.26&  0.16& 0.14\\
          &Separate words&  0.22&  0.13& 0.09\\
          &Template&  0.18&  unsupported  & 0.07\\
          &Multiple choice&  0.20&  0.09& 0.01\\
          \textbf{Spontaneous sentences}&None&  0.41&  0.25& 0.15\\
          &Animals&  0.39&  0.24& 0.13\\
          \textbf{Full spontaneous fragments}&None&  0.43&  0.89& 0.20\\
  &Animals& 0.42& 0.89&0.19\\
  && & &\\
  && & &\\
    \end{tabular}
\end{table}
 \vspace{-1.7cm}

\subsection{Responsiveness}
\begin{figure}
\captionsetup[subfigure]{justification=Centering}

    \begin{subfigure}[t]{0.45\textwidth}
        \includegraphics[width=\textwidth]{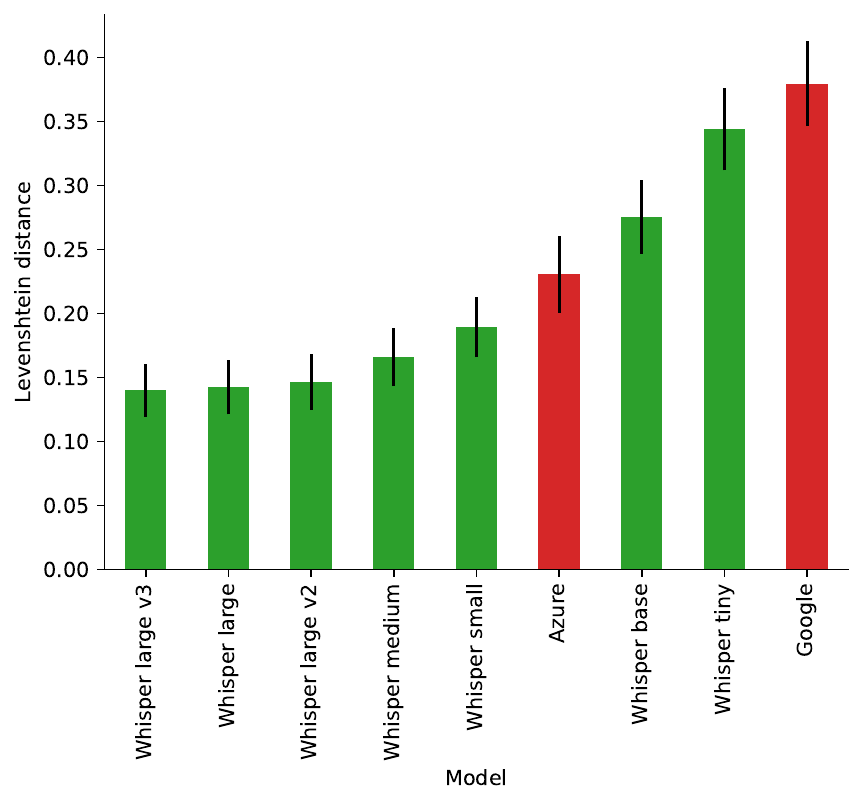}
        \caption{Performance of all current ASR engines (Whisper model versions in green, Google and Azure in red) calculated as Levenshtein distance between ground truth and transcription (lower is better).}
        \label{fig:Levenshtein_models}
    \end{subfigure}\hspace{\fill} 
    \begin{subfigure}[t]{0.45\textwidth}
        \includegraphics[width=\linewidth]{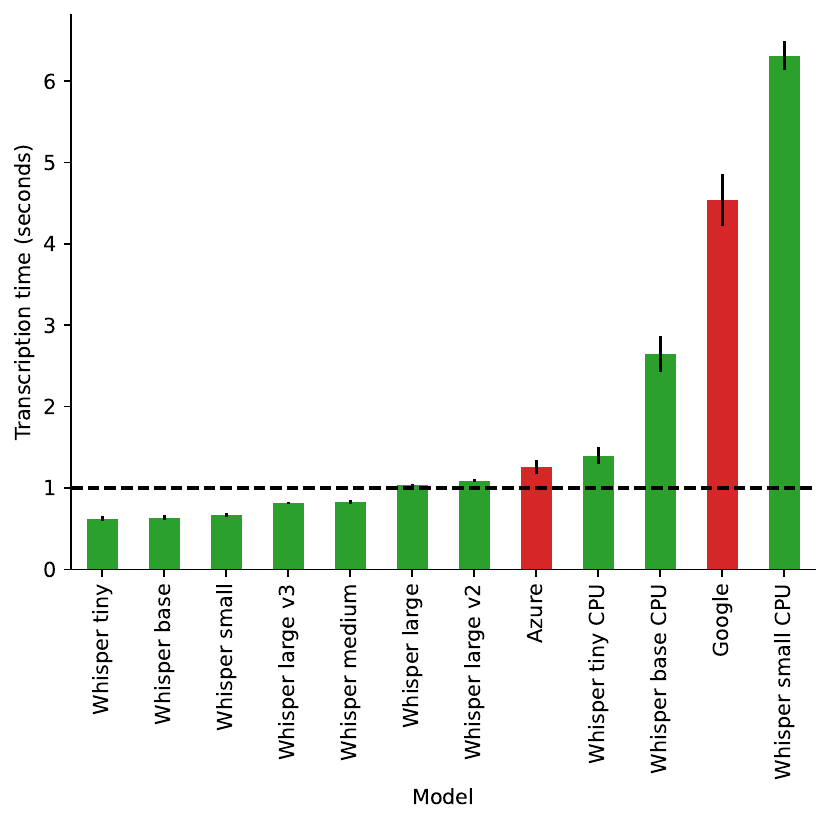}
        \caption{Average transcription time for a sentence. Whisper models (in green) were run on GPU unless CPU is specified. Dashed line shows maximum acceptable delay of 1000ms.}
        \label{fig:time}
    \end{subfigure}
    
    \bigskip 
    \begin{subfigure}[t]{0.45\textwidth}
        \includegraphics[width=\linewidth]{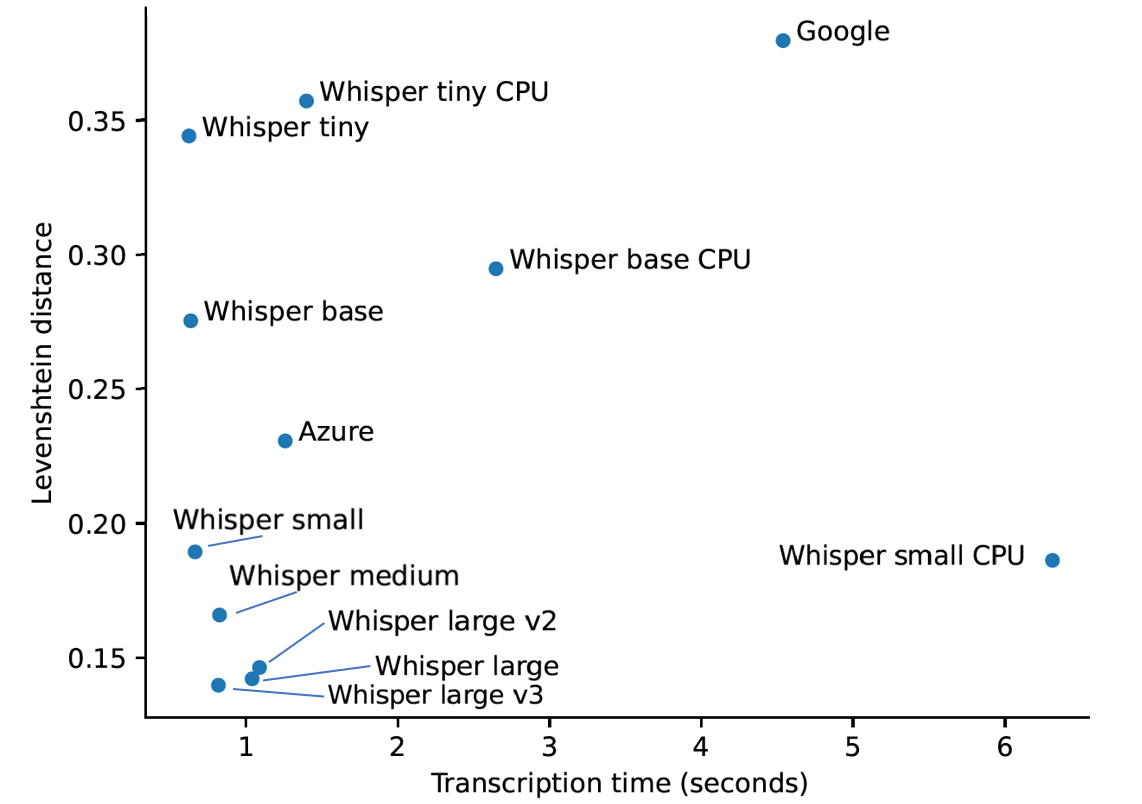}
        \caption{Transcription time vs. accuracy (lower is better). Whisper models were run on GPU unless CPU is specified. Ideal ASR systems would be in the lower left corner.}
        \label{fig:scatter}
    \end{subfigure}\hspace{\fill} 
    \begin{subfigure}[t]{0.45\textwidth}
        \includegraphics[width=\linewidth]{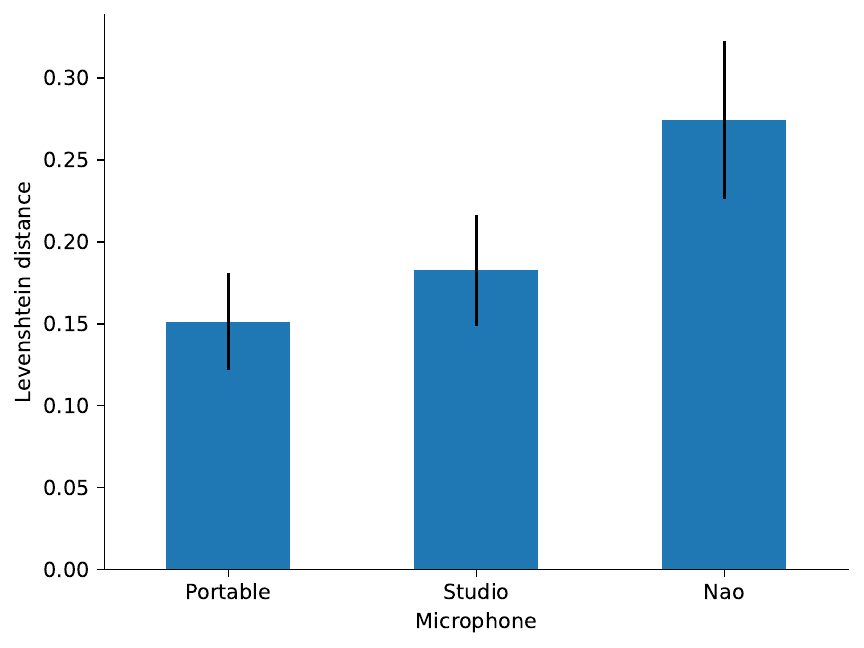}
        \caption{Performance of Azure, Google and best Whisper model when using different microphone types, calculated as Levenshtein distance (lower is better). Best results are obtained when using a microphone external from the robot.}
        \label{fig:microphone}
    \end{subfigure}

\caption{Graphs comparing the performance and transcription time of all current ASR engines and the impact of the microphone type on these metrics.}

\end{figure}

In Figure \ref{fig:time}, the average transcription times for short sentences (spontaneous speech and repeat sentences) are shown for Google, Azure, all of the Whisper models on GPU and the tiny, base and small Whisper models on CPU. The average transcription time for the Whisper medium and large models on CPU are respectively 17.5s and 30.5s, and were left out of the graph for readability. We visually mark the 1000ms line on the figure because, even though the mean response time in human conversation is 200ms, for spoken dialogue systems a delay of between 700 and 1000ms is deemed acceptable \cite{skantze2021turn}. From this data, it can be concluded that using a local model run on a GPU, instead of CPU or using an API, can greatly improve the responsiveness, until an acceptable level for spoken dialogue.

The Kruskal-Wallis test shows significant differences between the transcription time of the tested models ($p<0.001$), and post-hoc Dunn tests with Bonferroni correction show significant differences between all pairs of models, except for between Whisper tiny, Whisper base and Whisper small, between Whisper large v3 and Whisper medium, and between Whisper large, Whisper large v2 and Azure.

Figure \ref{fig:scatter} shows the relation between the models' average transcription time with their accuracy using the Levenshtein distance. To choose which model to use, both responsiveness and performance should be taken into account. Lower results are preferred for both, so models in the lower left corner of the scatter plot are ideal. As apparent in the figure, there is a trade-off between transcription time and transcription performance, so the choice should be made based on the specific application.


\subsection{Microphone}

Figure \ref{fig:microphone} shows the Levenshtein distance when using audio recorded by the three different microphones. Here, the transcriptions by Google, Azure and Whisper large v3 were used. When comparing the results of the internal Nao microphone with the portable and studio microphone, the Kruskal-Wallis test shows a significant difference between the groups, and Dunn's test with Bonferroni corrections as post-hoc analysis shows a significant difference between the Nao and portable microphone ($p < 0.001$) and the Nao and studio microphone ($p < 0.01$). There is no significant difference between the studio and portable microphone ($p=0.399$). In conclusion, the worst results are obtained when the microphone in the Nao robot is used, as there is a lot of added noise due to the closeness to the robot's motor and ventilation, but no difference is found between both external microphones.



\subsection{Energy consumption}

As concerns have been raised over the energy consumption and consequently carbon emissions of state-of-the-art machine learning \cite{lacoste2019quantifying}, we think it is valuable to consider these for ASR systems. We only have been able to measure Whisper's consumption. Per hour of transcribed data, the largest and best-performing model (large-v3) consumes $32.3 Wh$ and produces $7.7 gCO_{2}eq$. Note that this energy consumption refers to the transcription of many samples back-to-back, while real-time interactive systems would not be able to work as efficiently.

\section{Conclusion}

Based on our evaluation, we can make the following recommendations, updating or overriding those made in \cite{kennedy2017child}:

\begin{description}
    \item [Recognition performance.] The recognition performance has improved dramatically for state-of-the-art ASR, with the best models of 2024 showing over 60\% fewer transcription errors than in 2017. However, performance falters when input samples become much shorter or longer than a single sentence. Model priming is able to improve performance, especially in very structured settings, e.g. when a number of fixed options are expected. 
    Adult-like recognition is not available yet, but the semantic content of children's speech is now sufficiently transcribed to offer potential for robust spoken interaction. Small errors can often be caught by other components of the dialogue management system ---such as large language models--- resulting in a higher task accuracy than the reported accuracy of ASR systems.
    
    \item [Responsiveness.] The responsiveness of locally hosted models (in our case OpenAI's Whisper) is significantly better than that of cloud-based solutions, with sub-second results for some models. The network overhead and shared services of using cloud-based solutions are not optimal for real-time spoken interaction, and local models even outperform the cloud-based solutions in accuracy.
    \item [Impact of microphone.] Using an external microphone, as opposed to a microphone embedded in the robot, leads to a significantly improved recognition performance. Performance improves regardless of the quality of the microphone, as the robot's noise has a stronger effect on the speech recognition than the choice of microphone.
\end{description}



\begin{credits}
\subsubsection{\ackname} This research received funding from imec (Smart Education), the Flemish Government (AI Research Program) and the Horizon Europe VALAWAI project (grant agreement number 101070930). We are indebted to the authors of \cite{kennedy_2016_200495} and \cite{kennedy2017child} for making the recordings and transcriptions available.

\end{credits}

%
%
\bibliographystyle{splncs04}
\bibliography{references}
%




\end{document}